# Memristive Threshold Logic Circuit Design of Fast Moving Object Detection

Akshay Kumar Maan, Dinesh Sasi Kumar, Sherin Sugathan, and Alex Pappachen James

**Abstract**

Real-time detection of moving objects involves memorisation of features in the template image and their comparison with those in the test image. At high sampling rates, such techniques face the problems of high algorithmic complexity and component delays. We present a new resistive switching based threshold logic *cell* which encodes the pixels of a template image. The cell comprises a voltage divider circuit that programs the resistances of the *memristors* arranged in a single node threshold logic network and the output is encoded as a binary value using a CMOS inverter gate. When a test image is applied to the template-programmed cell, a mismatch in the respective pixels is seen as a change in the output voltage of the cell. The proposed cell when compared with CMOS equivalent implementation shows improved performance in area, leakage power, power dissipation and delay.

**Index Terms**

Threshold Logic, Resistance Networks, Memristors, Object Detection

## I. Introduction

In human brain, motion-detection is a low level visual activity and is the early form of visual intelligence. Inspired from the fact that human visual cortex is robust in object detection and tracking [1], [2], in this paper we attempt to mimic the functional mechanisms of the brain in a high speed programmable resistance VLSI circuits. The object feature detection and tracking is highly investigated problem in the last decade, taxonomy of which includes point tracking [3]–[5], kernel tracking [6], [7] and silhouette tracking [8], [9] and bio-inspired neuromorphic circuits based on analog computation and localized memory [10], [11]. Typically, in an algorithmic approach the desired object to be tracked uses a limited set of image sequences. Conventional visual object tracking solutions are limited by high computational complexity, inability to handle high resolution images, low frame sampling rates, and less flexible hardware architectures. The number of image sequences (or frames per second) that can be utilised for tracking varies from one algorithm to another.

Hardware implementations of object recognition algorithms are usually limited by the frame rate or by the area on chip. The FPGA (Field-Programmable Gate Array) based object detection implementation such as in [12] results in a maximum processing speed of 15 frames per second. The complexity and on-chip area is another concern such as presented in retina inspired design [13], and bio-inspired multi-level architecture [14]. Neuromorphic object detection cells [15] demand a considerable amount of chip space. However, the complexity of implementing neural interconnections certainly limits the scalability of such architectures [16], [17]. Resistance based neuromorphic models [18] approach the problem of object detection as a filtering mechanism resembling the human eye, that in so-far is limited by 50fps speed and large on-chip area. Optical flow estimation techniques [19] are also widely used in motion estimation that have a considerable amount of circuit level complexity. The winner-takes-all neural networks [20], sensor level processing [21] and analog neural networks [22] have found application in object detection with limited scalability.

In this paper, we advance towards a neuron inspired approach to moving object detection problem. We introduce a concept of pattern matching based on bilevel threshold logic cells formed in a network of programmable resistances. In contrast with existing techniques, we adopt a hardware oriented approach to object tracking problem, where in principle and practice very high frame rates and high resolution video images can be utilised. The proposed neuromorphic circuit can be seen as a hardware model for human brain in object detection and, an early stage of visual intelligence in VLSI.

Manuscript received December 12, 2013; revised April 14, 2014 and July 27, 2014; accepted September 17, 2014. This work was supported by the Enview Research and Development Laboratories, Trivandrum, India, under Grant 2013CV/03.

A. K. Maan is with the School of Information and Communication Technology, Griffith University, Nathan, QLD 4111, Australia (e-mail: akshay.maan@gmail.com).

D. S. Kumar and S. Sugathan are with the Enview Research and Development Laboratories, Trivandrum 695011, India (e-mail: apj@ieee.org; sherin2701@gmail.com).

A. P. James is with the Department Electrical and Electronic Engineering, Nazarbayev University, Astana 010001, Kazakhstan (e-mail: apj@ieee.org).

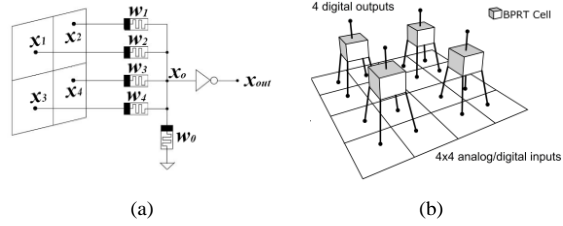

(a) (b)

Fig. 1: (a) Example of a BPRT cell that uses a 4 input pixel values. The example shows 4 input weights that are individually connected to separate inputs. (b) An architectural representation of the cell arrangement for a $4 \times 4$ pixels image; in this configuration, each cell uses 4 analog/digital inputs, has bi-valued weights, and implements image dimensionality reduction.

## II. BILEVEL PROGRAMMABLE RESISTANCE THRESHOLD LOGIC

Figure 1a shows the unit processing element in the network which we refer as a modular bilevel programmable resistance memristive threshold logic (BPRT) cell or in short *cell*. The collection of cells arranged referred to as a single layer forms the proposed network configuration as shown in Fig. 1b. The modular structure of network makes it easy to implement in hardware.

The parameters of proposed BPRT cells in Fig. 1b are weights $w_i (i = 1, 2, ....., N)$ and a constant weight $w_0$. The weights $w_1$ through $w_N$ are configured dynamically, using the reference inputs $x_i$. The weighted summation of inputs $x_i$ is denoted as $x_o$, while the threshold logic implemented using a logistic function results in a binary output $x_{out}$. The reference pixel inputs $x_i$ from a reference sample $X_r$ is used for setting $w_i$, and results in a binary output value of $1V$. When a test pixel $x_t$ is applied to the cell, the binary output value would change from $1V$ to $0V$ if dissimilar, but will remain at $1V$ if similar. The proposed cell is essentially a similarity calculator and performs a binary decision oriented pattern matching. It can be noted that, although in principle it may seem somewhat similar to some of the early hard-threshold neuron models, the proposed cell is conceptually and implementation wise different from the conventional threshold logic. In addition, the proposed logic can be realised as a hardware circuit with the weights translated as conductance (i.e. conductance of a memristor) for a potential divider and the threshold translated to an logic inverter as shown in Fig. 1a.

The Fig. 1b shows the block wise arrangement of inputs for processing by the BRT cells. Let $C$ be a cell in the arrangement with weight (conductance) values $w = \{w_1, w_2, ..., w_n\}$ and the corresponding inputs $x = \{x_1, x_2, ..., x_n\}$. Each of the weights $w_i$ ($1 \leq i \leq n$) of $C$ is set to either high $w_H$ or low $w_L$ as follows:

$$w_i = \begin{cases} w_H & \text{if } x_i > x_a \\ w_L & \text{otherwise} \end{cases} \quad (1)$$

where $x_a$ is the average of the normalized pixel values in the input frame and is calculated as $x_a = \frac{1}{n}\sum_{i=1}^{n} x_i$.

It is not practical or general to use Eq. (1) for setting bilevel weights as in practice it is difficult to implement an ideal inverter logic. An alternative approach to express the same idea of setting the $w_i$ to bilevel weight values using generalised logistic functions. The generalised logistic function with respect to weights can be expressed as: $w_i = \frac{w_H}{1+be^{cx_i}} + w_L$. The value of $c$ is obtained by equating this to $\frac{w_H}{2} + w_L$ and $x_i = x_a$. This results in $c = \frac{-1}{x_a}\log b$ and the equation for setting the weights becomes: $w_i = \frac{w_H}{1+be^{\frac{-x_i}{x_a}\log b}} + w_L$. The output voltage $x_0$ of the weighted part in the cell with $n$ inputs is given by:

$x_0 = \frac{\sum_{i=1}^{n} x_i w_i}{w_0 + \sum_{i=1}^{n} w_i}$. The threshold output of the cell can be conceptually expressed as:

$$x_{out} = \begin{cases} 1V & \text{if } x_0 < t_a \\ 0V & \text{otherwise} \end{cases} \quad (2)$$

where, $t_a$ is the logistic threshold. A more realistic that calculates $x_{out}$ can be represented as: $x_{out} = \frac{b_1 e^{\frac{-x_0}{t_a}\log b_1}}{1+b_1 e^{\frac{-x_0}{t_a}\log b_1}}$. Substituting the logistic equations of $x_o$, we obtain the $x_{out}$ as:

$$x_{out} = \frac{be^{-\beta \sum_{i=1}^{N} x_i w_i}}{1 + be^{-\beta \sum_{i=1}^{N} x_i w_i}} \quad (3)$$

where, $\beta = \frac{\log b}{t_a[w_0 + \sum_{i=1}^{N} w_i]}$. The Eq (3) and its logic counterpart Eq. (2) can be viewed as a transform for local similarity calculation, that compares a set of reference pixels from a template sample $X_r$ to that of the test pixels originating from a test sample $X_t$. The template sample pixels are memorised as conductance values $w_i$ which are set to bilevel values based on the intensity of reference input signals. The test sample does not set the weights, and hence the output of the Eq (3) would only change if the reference pixel intensities are significantly different from the test pixel intensities. Suppose in an image there are $M$ block of pixels that are to be compared. This means there will be $M$ cells, the global similarity $s_g$ between the test and reference sample image can be obtained by the summation of $x_{out}$ across the sample: $s_g = \frac{1}{M} \sum_{j=1}^{M} x_{out}(j)$.

*a) BRT cell working:* Figure 1a, shows the hardware model of a 4-input BRT cell in the network shown in Fig. 1a. In this cell the weights, $w_1, w_2, w_3$ and $w_4$, will be fixed to $w_H$ or $w_L$ as per the Eq. (1). For an illustration, consider that the values for $w_H$, $w_L$ and $w_0$ are conductance $\frac{1}{10}\mu S$, $\frac{1}{100}mS$ and $\frac{1}{50}mS$, which can be obtained by changing the memristance of a memristor. The threshold for the inverter has been set as $t_a = 0.5V$. As per the Eq. (3), if the output of the potential divider part ($x_0$) is greater than $t_a$ the output of the threshold will be $0V$; otherwise the output will be $1V$.

Now, take a scenario in which an image is 4×4 pixels and the average of the pixel values of the image is $0.6V$ which is $x_a$ of this network, that means all cells in the network will have the $x_a$ value $0.6V$. In this case the network contains 4 cells in which each cell will have four inputs as shown in Fig. 1a. Suppose for one cell, the input pixels voltage values are, $x_1 = 0.2V$, $x_2 = 0.3V$, $x_3 = 0.3V$, and $x_4 = 0.1V$. These voltage values are considered as the template pixel values. The weights $w_1, w_2, w_3$ and $w_4$ of the cell will be fixed to $w_L, w_L, w_L$ and $w_L$ by changing the memristance of the corresponding memristor as per the Eq (1). Then the voltage at node $x_0 = 0.15V$, which is smaller than the threshold value of the inverter $t_a = 0.5V$, so the output of the cell becomes $1V$. For training the cell, the memristor's weight need to be set using an additional circuit as given in the Fig. 2. This will avoid the unwanted resistance change in the circuit during the test phase. The switches $S_1$, $S_2$ and $S_3$ will switch as per the control signal $t_{tr}$, where $t_{tr}$ is high (switch connected) during training phase. The Opamp is designed using eight MOSFET's [23], where its area is $31.3\mu m^2$ and $16\mu W$ power dissipation and is used as a comparator is to enable the proper control of the selection inputs. This circuit can be used in serial or parallel to set the memristance of the memristors in training stage. The possibility of serially training the cells based on pixel values using array switched connections is shown in Fig. 2b.

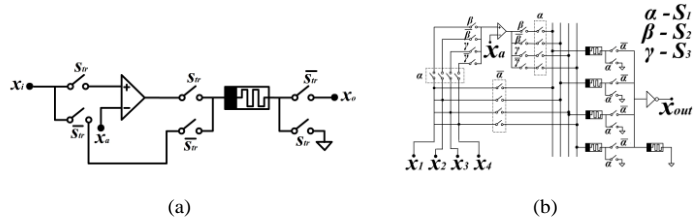

(a)  (b)

Fig. 2: (a) shows the training circuit required to set memristance to high or low value during training stage taking into account the input voltage of the reference pixel, and (b) shows an array realisation of the circuit for network.

To illustrate the change detection in cell, suppose that there is a significant change in input values (change in pixel intensity) when the next frame input is applied, say $x_1 = 0.9V$, $x_2 = 0.9V$, $x_3 = 0.8V$, and $x_4 = 1V$. In the test stage, the weights remain at the same values as that were set using template image pixels. The voltage at the node $x_0$ will change from $0.15V$ to $0.6V$, which is higher than the threshold value of the inverter and hence the output of the cell will change to $0V$. It is to be noted that the change occurred here is a change from dark pixel to light pixel and a change is detected by the cell. However, the change will not get registered if there are only slight variations in the inputs. For example, if the new input values are $0.3V$, $0.3V$, $0.3V$ and $0.1V$ which reflects a small change from the template pixels, then the output of the potential divider becomes $0.16V$, which is obviously insufficient for the inverter to change its output. In this sense, it can been said that the output of the cell will only change during significant changes in the input. It is from the test pixel changes that we observe that the cell will detect the changes only from black to white and not the reverse. Table I shows the four input cell properties.

TABLE I: Area, power dissipation, leakage power and delay of a four input cell, where $w_H$, $w_L$ and $w_0$ are $\frac{1}{10}\mu S$, $\frac{1}{100}mS$ and $\frac{1}{50}mS$

| Area ($\mu m^2$) | Power dissipation ($\mu W$) | Leakage Power (pW) | Operational Speed (MHz) |
|---|---|---|---|
| 9.66 | 12.30 | 12.25 | 100 |

Figure 3 shows the timing behaviour of the proposed 2-input cell having $x_1$ and $x_2$ as the input pixels and $x_{out}$ the output pixel, where the cell is trained to detect dark pixels ($0V, 0V$), with both the weights set to $w_L = \frac{1}{100}mS$. The pixel values change in $10ns$ and the behaviour is shown for a period of $40ns$. From the waveform we can see that the cell is able to detect the change from ($0V, 0V$) to ($1V, 1V$), but it fails to detect the changes from ($0V, 0V$) to ($0V, 1V$), or ($1V, 0V$). This is because the cell is required to detect the changes only if there is a sufficient change in the group of pixels within the image and inherits the ability to discard the random pixel changes from the sensor noise.

The proposed work is compared with the benchmark neuromorphic learning chip [14] that target object detection and tracking. In order to do the comparison we have done simulation using similar technology scaling of learning chip [14]. The simulation is performed in SPICE (Simulation Program for Integrated Circuit Emphasis) using feature size of 0.35 µm TSMC (Taiwan Semiconductor Manufacturing Company) process BSIM (Berkeley Short-channel IGFET Model) models and HP memristor model [24]. The results in Table II, show a clear advantage for the proposed method, in terms of performance when compared with the neuromorphic learning chip[1] [14].

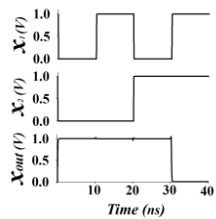

Fig. 3: Waveform showing the response of a 2-input cell, which has trained to (0V ,0V ) input template pixels. We can see that the cell is able to detect the change from (0V ,0V ) to (1V ,1V ) by changing $x_{out}$ to 0V from 1V .

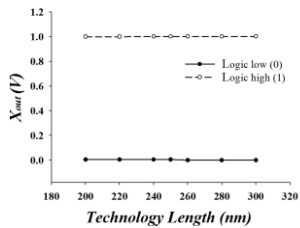

Fig. 4: Effect of change in technology length on a 4 input cell

TABLE II: Performance Comparison between a BRT network of 32 Neuron Cell with the Learning Chip in [14]

| Parameters | BRT network | Learning Chip [14] |
|---|---|---|
| Area ($\mu m^2$) | 309 | 1936 |
| Power dissipation (W) | 396µ | 60m |

Throughout this paper, we use the non-ideal resistive switching model of memristor reported in [24] for our study with an area of 10nm×10nm and resistances in the range of [$10^6$, $10^{12}$]Ω, while CMOS circuits use 0.25µm TSMC technology to reflect the practical standard silicon technologies. The estimation of the area is done at the device level only. The reported area of a network were estimated based on the area of a macro model used. Figure 4 shows the effect of change in technology length on the output of a 4-input cell. From the figure we can see that a ± 50nm change in the technology length does not introduce any effective change in the cell response both in case of logic high (1) and logic low (0). Since the physical design of the memristor is not available to simulate, practically we will not be able to check the scalability issues and the response plotted in the Fig. 4 is the simulated results using non-ideal memristor models in SPICE, and realistic CMOS process models.

*b) Fast Object Detection:* In object detection, a template image frame is chosen from the video input and subsequent test image frames are compared with the template frame. We require two parallel BPRT networks in order to detect both light to dark and dark to light pixel intensity changes in the object. This method is shown in the Fig. 5a where each BRT network is shown as a separate modules. Module 1 of the BPRT network is meant for the detection of dark to light pixel intensity changes and module 2 for the light to dark intensity change detection. The resistor weights in the module 1 will be set using the template image pixel and that of module 2 with the inverted template image pixels during the training phase of the network. Each cell in the modules contains 4-inputs, to which 4 input pixel voltages from input images are applied as shown in the Fig. 5a. The resistor values (weights) of each cell is fixed based on the input pixel values, such that each cell in the module is trained to detect any sufficient changes occurring in the input set of pixels. The output of the modules is of size equal to the number of cells as shown in Fig. 5a. These outputs have a value of 1V when the template pixels are applied as the inputs. The output of two modules will be combined pixel-wise using AND gates in order to merge similarity detection results of both dark to white and white to dark changes.

The processing speed of few megahertz would mean that more than $10^6$ images can be processed in 1 second. Even when delay of the training elements are taken into account this would still be more than sufficient to deal with high frame rates of a camera. In practical terms, as the frame rates of the camera are still limited to a few tens to hundreds of frames per second, the proposed object detection would not have issues with continuous update of the memristors. However, in future when extremely higher frame rates are required, the improvement in operating speeds would be required, that is largely dependent on the operation speed of the operational amplifiers and CMOS invertors in the circuit.

---

[1]The results compared are specific to the cell level. The board level performance is not compared as they would be different in application and need.

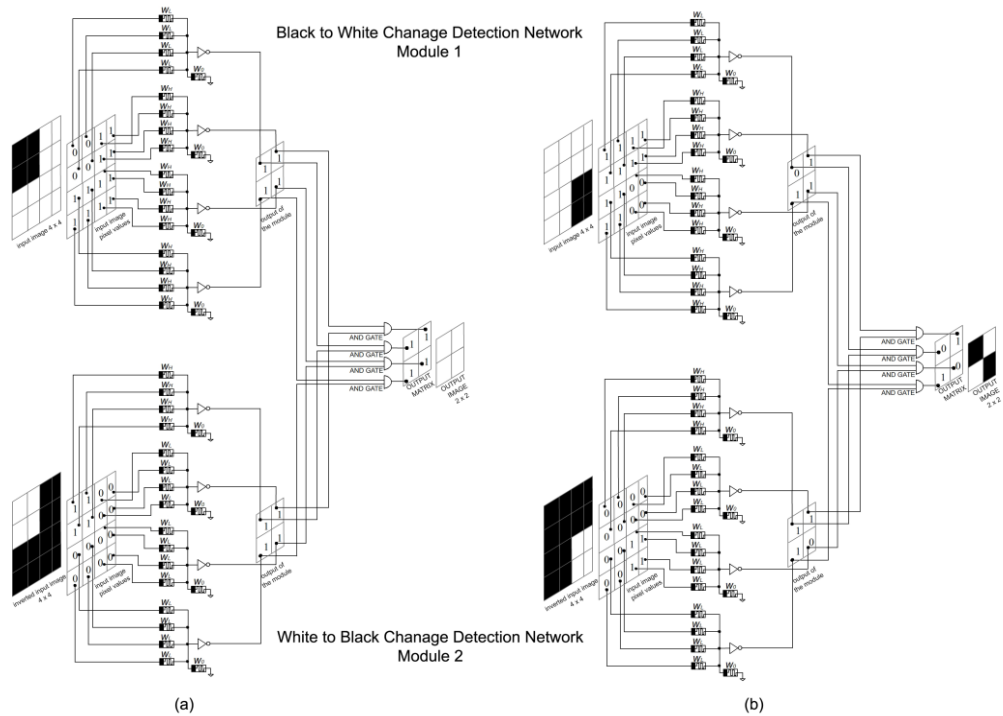

Fig. 5: (a) Training of a BRT network for object detection with an input image of size 4 x 4 pixels. Module 1 has been trained with original input image and Module 2 has been trained with the inverted input image. (b) A new image is passed through the trained network resulting in change detection at the output.

When the test frames are applied, if there is any sufficient change in the set of pixel values, the output of that particular cell will change from 1V to 0V. This behaviour of the cell is shown in the Fig. 5b. Here the image is different from the trained template image. It has both dark to white and white to dark changes and both this changes are detected by the corresponding modules. In addition, it can be seen that the values of the output in the detection of moving object has been recorded by changing the corresponding cell output to 0V. When combining the module output values, we are able to detect the total changes in the image. This is shown in the output image as two dark pixels. The unchanged values are the static background and the changes detected are the objects. From, Fig. 5a and Fig. 5b it is clear that since the size of cell is 2 × 2, the size of the output image will be reduced to $\frac{N}{2} \times \frac{N}{2}$, where N × N is the input image size.

TABLE III: COMPARISON OF PERFORMANCE PARAMETERS FOR TWO DIFFERENT BRT NETWORKS.

| Performance parameter | Value (4×4 pixels) | Value (352×288 pixels) |
| --- | --- | --- |
| Area | 193$\mu m^2$ | 1.22$mm^2$ |
| Power dissipation | 103$\mu W$ | 0.65W |
| Leakage power | 245pW | 1.55$\mu W$ |
| Operational speed | 100MHz | 100MHz |

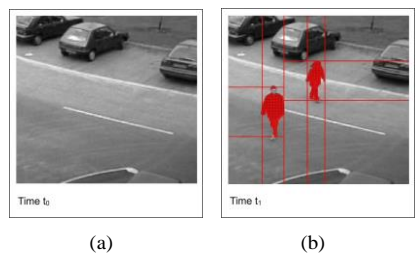

Fig. 6: The illustration showing (a) the template image pixels and (b) the test image with object pixels detected.

TABLE IV: THE PERFORMANCE RESULTS OF THE PROPOSED TECHNIQUE

| Database object label | Sensitivity (%) | Specificity (%) | False positive (%) | False negative (%) | Youden's Index | Precision (%) | Positive likelihood | Negative likelihood | F-measure (%) | Accuracy (%) |
|---|---|---|---|---|---|---|---|---|---|---|
| ViSOR | 92.3 | 96.8 | 3.2 | 7.7 | 0.8906 | 99 | 28.6 | 0.1 | 95.5 | 93.3 |
| PETS 2000 | 100 | 100 | 0 | 0 | 1 | 100 | ∞ | 0 | 100 | 100 |
| CDVP | 89.8 | 92.4 | 7.6 | 10.2 | 0.8218 | 92.2 | 11.8 | 0.1 | 91 | 91.1 |
| SPEVI | 98.3 | 100 | 0 | 1.7 | 0.9833 | 100 | ∞ | 0 | 99.2 | 99.2 |

Figure 6 shows the illustrative result of moving object detection using the proposed cells. The simulation shown in Fig. 6 has been done using an array of 4-input cells and with an input frame size of 352 × 288 pixels. The values of the performance parameters (for the input frame size of 352 × 288 pixels) are given in Table III. The network gives an output at a reduced dimension of 176 × 144 pixels. We used the macro-model cells in Fig. 1 to simulate the object detection in 352 × 288 pixels. The pixels were processed within SPICE simulation environment through reading image frames from the benchmark object detection databases (ViSOR, PETS 2000, CDVP, and SPEVI databases). The binary changes detected in the output are then plotted back on the original image for display purposes. The template image at time $t_0$ (Fig. 6a) is the first frame in the video and does not have any moving objects, while at in a subsequent frame at $t_1$ objects (Fig. 6b) enter the scene and are detected as shown by the red colored pixels. The results are reported using ViSOR, PETS 2000, CDVP, and SPEVI databases. The performance values on object tracking tests across the databases result in an average F-score of 96.4%, accuracy of 95.9%, Youden's index of 0.924, sensitivity of 95.6%, and specificity of 97.3%. Table IV shows the breakup of these results for the databases. Figure 7 shows the receiver operator characteristic of the object detection on the four databases using our proposed cell. The plot has been generated based on the amount of object blobs that have been correctly classified. Although objects are detected in all the frames, those that are difficult to recognise due to overlap with other objects in the same frame are considered to be not detected to make it realisable in practical object tracking tasks.

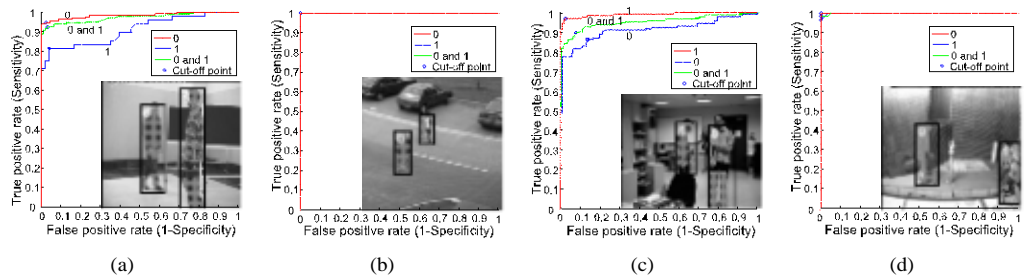

(a)  (b)  (c)  (d)

Fig. 7: A graphical illustration of performance of the proposed method in realistic object detection situations: (a), (b), (c), and (d) show the receiver operator characteristics, along with a sample image frame with obtained detections for ViSOR, PETS 2000, CDVP and SPEVI databases. The boundaries shown are for illustration purpose.

There exists low-power programmable circuits mimicking neural behaviour [25], [26] that can be used for object detection applications. However, any objective comparison between the proposed cell and those of the like in neuron-synapse in [25] is difficult with different technologies without reporting the power and area specific at the cell level. The classification part in similar works assumes the knowledge of gallery that can be understood as a offline matching method. In contrast, the proposed method produces templates on the fly (online processing) and the detection of objects are done on the incoming images without storing the templates, rather is encoded in the network of the cells.

## III. CONCLUSION

A programmable memristive threshold logic circuit for moving object detection was presented in this paper. An important advantage of the method discussed is that it can support an implementation targeted at high speed imaging. The hardware nature of the cell architecture makes it an alternative to algorithmic approaches and highly attractive to the practical utilisation of high frame rates that could lead to near-continuous real-time object tracking, and even surpass human object tracking ability. The presented logic provides a framework to implement brain like logic in a memory and learning driven detection of multiple objects. In terms of hardware implementation, this needs memristor like devices that can do large number of cycles, smaller area and long data retention time which may be made possible with emerging memory technologies. A truly reconfigurable and dynamic cell for multiple object detection, tracking and recognition in high resolution and high speed videos is a topic for further research, that may have a significant impact on the way object detection in real-time is implemented in hardware.